\documentclass{llncs} 
\usepackage{rotating} 
\usepackage{amsmath}
\usepackage{amssymb}

\usepackage[UKenglish]{babel} 
\newcommand\fl{\mathbin{\rightarrow}}
\newcommand\ttt{\mathsf{t}}
\newcommand\eee{\mathsf{e}}
\renewcommand\ggg{\mathsf{g}}
\newcommand\systF{\ensuremath{\mathsf{F}}}
\newcommand{\editout}[1]{}

\newcommand\ltyn{\ensuremath{\Lambda Ty_n}}
\newcommand\ma[1]{\textit{``#1''}}
\newcommand{\ourskip}{\medskip}

\usepackage{linguex}

\title{Plurals: individuals and sets\\ in a richly  typed semantics
\thanks{Work supported by ANR project Polymnie}
}
\titlerunning{Plurals: individuals and sets in a richly  typed semantics} 
\author{Bruno Mery \and Richard Moot \and 
Christian Retor{\'e}
\thanks{On CNRS sabbatical at IRIT (Toulouse) from September 2012 to August 2013}  
}
\institute{
LaBRI-CNRS \& Universit{\'e} de Bordeaux\\
}
\authorrunning{B. Mery, R. Moot, Ch. Retor{\'e}}

\begin{document}

\maketitle

\begin{abstract} We developed a type-theoretical framework for natural language semantics that, in addition to the usual Montagovian treatment of compositional semantics, includes a treatment of some phenomena of lexical semantic: coercions, meaning, transfers, (in)felicitous co-predication.  In this setting we see how the various readings of plurals (collective, distributive, coverings,\ldots) can be modelled. 
\end{abstract}


\section{Introduction}



In this paper, we incorporate the treatment of plurals into the
unified type-theoretical framework of \cite{BMRjolli,MPR2011taln},
using polymorphism and type-coercions to account for the basic facts
of plurals.

We must warn the reader that the purpose of the present paper is neither to assert acceptability judgements nor to 
to provide linguistic criteria that  
predict the correctness of readings. Our paper provides 
a formal and computational account of linguistic analyses and theories
of plurals, which is fully compatible with the type theoretical
framework in which we studied other lexical or compositional
phenomena.  As such, a prototype  implementation has been
incorporated in our syntactical and semantical parser Grail \cite{moot10grail}.

The rest of this paper is structured as follows. Section~\ref{sec:plurals}
will give a brief overview of the problems a theory of plurals faces
and some of the solutions proposed for treating
them. Section~\ref{sec:tyn} introduces our type-theoretic framework
$\ltyn$ and compares it to other approaches. Section~\ref{sec:tynpl}
introduces the basic elements of our theory of plurals, which will be
developed with a fully work out lexicon and several examples in the
following section.

\section{Plurals in formal semantics: known difficulties, typical examples and classical models} 
\label{sec:plurals}

\subsection{Plural readings} 

Plurals are one of the most ubiquitous of semantic phenomena. Though
not explicitly treated by Montague himself, a large body of research
has been developed since then and we cannot hope to treat all the
subtleties in a single article --- the reader is referred to \cite{link,plurals} and
references therein for further discussion.  However, we will present several of
the basic facts and propose a simple yet detailed Montagovian
treatment of them, doing so in a way which makes both the combinatorics and the
available readings very clear.

Some predicates (typical examples include verb phrases like ``meet'',
``gather'' and ``elect a president'', but also prepositions like ``between'') require their
argument to be a group. These predicates are called
\emph{collective}. Below, some typical examples are listed.

\ex. \a. *Jimi met.
\b. Jimi and Dusty met. (unambiguous).
\b. *The student met. 
\b. The students met. (unambiguous, one meeting)
\b. The committee met. (unambiguous, one meeting)
\b. The committees met. (ambiguous: one big meeting, one
meeting per committee, several meetings invoking several committees)

In these cases, the subject can either be a conjunction of noun
phrases, a plural noun phrase or a singular group-denoting noun phrase
(typical nouns of this sort are ``committee'', ``orchestra'' and
``police''; in English these nouns require plural subject-verb
agreement).

Contrasting with the collective predicates 
 are the \emph{distributive} predicates. When they
apply to a set of individuals, we can infer that the predicate applies
to each of these. Examples include verbs like ``walk'', ``sleep'' and ``be hungry''.

Many other predicates, like ``record a song'', ``write a report'' or ``lift a piano'' are
ambiguous in that they accept both a group reading (which can be made
explicit by adding the adverb ``together'') and a distributive reading
(which can be made explict by adding the adverb ``each''). Thus
``recorded a song together'' talks about a single recording whereas
``recoded a song each'' talks about several recordings. The predicate
``record a song'' in and of itself only states that each of the
subjects participated in (at least) one recording. We call such
readings \emph{covering} readings.
The covering reading has both the
collective and the distributive reading as a special case.

We will only mention the so-called \emph{cumulative} readings \cite{scha81} in
passing: they include examples like ``Two composers wrote nine symphonies''
 which has --- in addition to the other readings --- a reading with a total of 18 symphonies. 

\subsection{Plurals in existing semantic theories}

Despite philosophical and technological differences, most of the work
on plurals in current theories of formal semantics can be traced
back in some way or another to the ideas and formalization of Link
\cite{link}, with many variations on the basic theme. In Link's treatment, groups exist at the same level as individuals
and we can form new groups out of individuals and groups by means
of a lattice-theoretic join operation, written $\oplus$ ---
mathematically, we have a
complete atomic join-semilattice without a bottom element. Groups and
individuals are related by the \emph{individual part} relation
$\leq_i$, defined as $a \leq_i b\ \equiv_{\textit{def}}\ a\oplus b = b$; in
other words $a\leq_i b$ is true iff $a$ contains a subset of the
individual members of $b$.

Unlike later treatments of plurals such as \cite{KR93}, Link does not distinguish atomic individuals from group
individuals but instead defines atomic individuals by means of a
unary predicate $\emph{atom}$, which is true for an entity $x$
whenever its only individual part is itself (ie.\ the set of atoms are those $a$ such that for all $b$, $b
\leq_i a$ implies $b = a$).

For arbitrary unary predicates $P$, we have a corresponding
distributive predicate $P^*$ which is true if $P$ holds for all atomic
subparts of its argument. For example, $\textit{child}^*(x)$ is true if
$x$ is a non-atomic group of children.\footnote{Technically, in
  \cite{link}, Link
  introduces two versions of this predicate: one which presupposes its
  argument is non-atomic and another which does not. Here, we will
  only consider the version which takes non-atomic individuals as an argument.}

We can therefore distinguish between 

\ex. \label{expl}
\a. $\exists x. \textit{child}(x)
\wedge \textit{sleep}(x)$ (a child slept),
\b. $\exists x. \textit{child}^*(x) \wedge
\textit{sleep}^*(x)$ (children slept), and
\b.\label{ex:raft} $\exists
x. \textit{child}^*(x) \wedge \textit{build-raft}(x)$ (children built a
raft, as a group).

\noindent where \ref{ex:raft} is shorthand for
$\exists x. \forall y. ((y \leq_i x \wedge \textit{atom}(y)) \Rightarrow
  \textit{child}(y)) \wedge \textit{build-raft}(x)$, with
  $\textit{atom}(y)$ in turn an abbreviation for $\forall z. z \leq_i y
    \Rightarrow z = y$.
\section{A Montagovian generative lexicon \ltyn}  
\label{sec:tyn}

\subsection{Lexical semantics in a compositional framework} 

The semantical analysis of natural language, computational and compositional,  consists in mapping 
any sentence to its logical form that is a  logical formula  which depicts its meaning. To do so, the lexicon provides each word 
with a typed $\lambda$-term. 
By induction the parse tree, we obtain  a  $\lambda$-term of type $\ttt$, i.e. a formula 
See eg.\  \cite[Chapter 3]{MootRetore2012lcg}. 
Usually there is just one base type for entities  $\eee$, while we shall use many of them $\eee_i$
--- base types can be taken to be the classifiers, for languages with classifiers like Japanese, Chinese, sign languages,...  \cite{MeryRetore2013nlpcs} With many base types the 
application of a predicate to an argument may only happen when it makes sense (\ma{The chair barks.} and  
\ma{Their five is running.}
are blocked 
by saying that \ma{barks} and \ma{is running}  
apply to individuals of type \ma{animal}. 
But some flexibility is needed. For instance, in a football match, \ma{their five} may be understood as a player, who can run. 


\subsection{\ltyn: lexicalist view} 

To handle those meaning transfers and coercions,
we reorganised the lexicon, starting with \cite{BMRjolli}. 
Each word is endowed with a main $\lambda$-term, the ``usual one'' which specifies the argument structure of the word, by using refined types: \ma{runs: $\lambda x^{animal} \underline{run}(x)$} only applies to \ma{animal} individuals. In addition, the lexicon may endow each word with a finite number of $\lambda$-terms (possibly none) that implement meaning transfers or coercions. For instance a \ma{town} may be turned into an \ma{institution},  
a geographic \ma{place}, 
or a football \ma{club}  
by the optional $\lambda$-terms  \ma{$f_i$: town$\fl$institution},  \ma{$f_p$: town $\fl$ place} and  \ma{$f_c$: town $\fl$  club} 
--- $\lambda$-terms may be more complex than simple constants. 
Thus,  a sentence like \ma{Liverpool is a large harbour and decided to build new docks.} can be properly analysed. Some meaning transfers, like $f_c$,  are declared to be \emph{rigid} in the lexicon. Rigidity prohibits the simultaneous use of other meaning transfers. For instance, the rigidity of $f_c$ properly blocks  \ma{* Liverpool defeated Chelsea and decided to build new docks.}. 

Actually,  we do not use the simply typed $\lambda$ calculus but 
second order $\lambda$ calculus, Girard's system $\systF$ (1971),
which factorises operation that act uniformly on families of types and terms
--- cf. figure \ref{systemF} or \cite{Girard2011blindspot}. 
The formulae i.e. the terms of type $\ttt$ that represent meanings are
many sorted higher order formulae. 
Because of polymorphism,  quantifiers  are given a single type, 
$\forall,\exists:\Pi \alpha. (\alpha \fl \ttt) \fl \ttt$
as well as Hilbert's operators for generics $\epsilon,\tau: \Lambda \alpha.\ (\alpha \fl \ttt) \fl \alpha$  \cite{Retore2012rlv,Retore2013taln}. A single term expresses copredication over the different facets of the same object, a polymorphic \ma{and}: 
$\Lambda \alpha \Lambda \beta
\lambda P^{\alpha \fl \ttt} \lambda Q^{\beta\fl \ttt} 
 \Lambda \xi \lambda x^\xi 
 \lambda f^{\xi\fl\alpha} \lambda g^{\xi\fl\beta}.\ 
(\land^{\ttt\fl\ttt\fl\ttt} \ (P \ (f \ x)) (Q \ (g \  x)))$ 

\begin{figure}[t]
\small 
\begin{tabular}{p{0.31\textwidth}p{0.66\textwidth}}
Types: \begin{itemize}
\item 
$\ttt$ (prop) 
\item 
many  entity types $\eee_i$ 
\item 
type variables $\alpha,\beta,...$ 
\item 
$\Pi \alpha.\ T$ 
\item  
$T_1\fl T_2$ 
\end{itemize}
& Terms 
\begin{itemize} 
\item Constants and variables for each type
\item 
$(f^{T\fl U} a^T):U$ 
\item 
$(\lambda x^T.\ u^U):T\fl U$  
\item $t^{(\Lambda \alpha.\ T)} \{U\}:T[U/\alpha]$
\item $\Lambda \alpha. u^T:\Pi\alpha. T$ --- no free $\alpha$ in a free variable of $u$.
\end{itemize}
\end{tabular} 
The reduction is defined as follows: 
\begin{itemize} 
\item $(\Lambda \alpha. \tau) \{U\}$  reduces to $\tau[U/\alpha]$ (remember that $\alpha$ and $U$ are types). 
\item $(\lambda x. \tau) u$ reduces to $\tau[u/x]$ (usual reduction). 
\end{itemize} 
\caption{Terms and types of system \systF}
\label{systemF} 
\end{figure} 

\subsection{Relation to other models} 

Our model bears some resemblance with other type-theoretical approaches to natural language semantics, in particular with the works by Asher and Pustejovky \cite{AP01} that reached its apogee in Asher's book \cite{asher-webofwords}, and also with the subsequent work by Asher, Bekki, Luo and Chatzikyriadikis 
\cite{AsherLuo2012sub,Luo2012lacl,Luo2012cslp,BekkiAsher2012lenls}
--- Cooper also uses type theory for modelling natural language semantics, but in quite a different perspective, more related to feature logics \cite{Cooper2011lacl}. All these models are type driven. 

In the models of the first group, such as \cite{asher-webofwords}, the system at work is the simply typed lambda calculus enriched with rules issued from 
category theory, whose compatibility is assumed. It was mainly designed to handle objects that combine several senses and facets and copredications over these senses, which can be felicitous or not. The possible benefits of this extension for classical phenomena in compositional semantics has not been much investigated. 

The models of the second group, such as \cite{Luo2012lacl}, use variants of Martin-L\"of type theory, with at least two levels,  types and kinds, and predicative quantification over types: $\forall\alpha\in CN$ (where $CN$ is the kind of common nouns). 
Some questions of formal semantics like coordinations and quantification have been discussed in this setting. 

The main difference between our model and these two is that they are \emph{type driven} while ours is 
\emph{word driven}, thus allowing coercion for a word and not for another 
word with the same type --- for instance \emph{classe} (French) is a
set of pupils but can also mean classroom while \emph{promotion} (French), also a set of pupils, cannot mean a room. 
Recent work by Luo,  \cite{Luo2012lacl},
 is a bit closer to ours since common nouns are  viewed as \emph{types} (hence for them type driven or word driven make no difference) but then it prevents 
two common nouns to have to be declared as being in the same sort. 
Regarding other part-of-speech, all these  models are completely type driven, hence less adequate to model language idiosyncrasies. 

A related difference is that when composing two phrases, our model
uses coercions issued from both phrases to allow readings that would otherwise be impossible. 

A methodological difference is that we give a unified framework for both compositional semantics and 
lexical semantics and pragmatics which properly accounts for many lexical phenomena like coercions and copredications \cite{BMRjolli}, fictive motion \cite{MPR2011taln}, deverbals \cite{RealRetore2013jolli} as well as compositional phenomena like a revisited view of (generalised) quantification and determiners \cite{Retore2012rlv,Retore2013taln}, or on plurals  in this very paper. 
The fact that  the phenomena we study are all fully formalised in a unique logical framework which is known to be sound guarantees that the model can simultaneously account for all these phenomena. 

The ultimate advantage of our  unified view is that being completely
formalised within a single  logical system, we can implement the computational
analyses as soon as we model them. We have
currently implemented a prototype version of the ideas in this paper
in the Grail parser \cite{moot10grail}.

\section{Sets, predicates and types} 
\label{sec:tynpl}

\subsection{Types are not sets} 
\label{setsandtypes} 

A close look at the Montagovian setting draws a distinction between two logics: 

\begin{description} 
\item[Logic/calculus for meaning assembly] (a.k.a glue logic, metalogic,...) In the standard case, this is simply typed $\lambda$-calculus with two base types $\eee$ and $\ttt$ — these terms are the proof in intituitionistic propositional logic. \emph{Here the base types will be $\eee_i$ sorts of the many sorted logic, with a particular type for groups, $\ggg$ and one for propositions, Montague's $\ttt$.} 
\item[Logic/language for semantic representations, target logic] In the standard case that is higher-order predicate logic or first-order logic, using reification.  \emph{Here this logic will be many sorted higher order logic.} 
\end{description}

Given this context, we would like to insist that types are not usual sets. For instance disjunctive types and complement types are problematic, or at least quite different from their set-theoretical counterpart and there is no intersection: 
\begin{itemize}
\item Although there is a negation $\lnot A=A\fl\bot$, negation does not implement set theoretic complements. Indeed, there cannot be both terms of type $A$ and terms of type $A\fl
  \bot$. One of these two types must be empty. 
When considering the associated predicate $\tilde A:\eee\fl\ttt$, there is a complement: 
$\lnot^{\ttt\fl\ttt} \tilde A:\eee\fl\ttt$. 
\item Disjunctive types are often  left out of the type system, since to obtain normal form one needs to consider unpleasant commutative conversions Furthermore, the terms in a disjunctive 
type $A_1\lor A_2$ are not  the unions of terms of type $A_1$ and of terms of type $A_2$, but pairs $\langle i, t\rangle$ of 
an integer $i$ and a term of type $A_i$, hence they never contain any common term to types $A_1$ and $A_2$
--- even if one pure term is common to both, as a proof of $A_1\cup A_2$, it is a different typed term, and the integer in front event prevent to have proofs of $A$ as proof of $A\cup A$. 
\item There is no intersection of types. A conjunctive type $A_1\land
  A_2$  contains pairs of terms of respective types $A_1$ and $A_2$, but  no common term to $A_1$ and $A_2$. 
\end{itemize}

Nevertheless some types are closer to sets, namely \emph{data types}, see e.g. \cite{Girard2011blindspot}, 
defined from constants and operators like, integers, lists of objects of type $A$,  finite trees 
with or without labels etc. A data type $T$  resembles a set in the sense that there is a bijective correspondence 
between normal 
terms of type $T$  and 
data of type $T$. . 
Data types can either be defined internally as in plain \systF\ 
or by constants and operators with specific reduction rules 
as in G\"odel's system \textsf{T} or Martin L\"of type theory.

The easiest account of a set in our Montagovian setting is a predicate which is of type $X\fl \ttt$. 
In order to view the sorts 
$\eee_i$ as sets, for each of them we have 
 predicate $\widehat \eee_i(\_):\eee\fl\ttt$ that encodes the property 
of being of type $\eee_i$. The choice of the domain of the predicate here $\eee$, the sort into which base types $\eee_i$ maps,  is rather free, since the domain of a predicate can be both restricted (as a function can) and extended (by saying it is false elsewhere).  
Once we have the base predicates, with their types, and the predicate associated with base types, plus our system of coercions, that include ontological inclusions, we can then define sets as one usually does. For instance, black applies to physical objects, cats may be viewed as physical objects (meaning transfer), hence \ma{the black cats in the garden} is a well defined set. 
This is the way we modelled, predicates, determiners and quantifiers. \cite{Retore2013taln}

\subsection{Operators for handling sets} 
\label{sec:operators}

We can define operators that represent operations on sets, that can be assimilated with predicates, 
of type $\eee\fl\ttt$. As integers can be defined, we shall have a
constant $|\_|$ for counting elements satisfying a
predicate. Inclusion is easily defined as a function that maps pairs
of predicates to a proposition: $\subseteq: \Lambda \alpha \lambda
P^{\alpha\fl \ttt} \lambda Q^{\alpha\fl\ttt}\lambda x^{\alpha} P(x) \implies Q(x)$ --- $\alpha$ usually is $\eee$ but this is not mandatory. 
The logical operators such as $\land$ and $\lor$ between predicates correspond respectively to union and intersection of sets of terms,  as expected (while operations on types like conjunction and disjunction do not, cf.\ Section~\ref{setsandtypes}). 

We will also consider one entity type for groups called $\ggg$, and a function \emph{member} that relates a group with its members, and which doing so turns a group in to a predicate: 
$member:\ggg\fl\eee\fl\ttt$ --- or equivalently, says whether an
entity satisfies a predicate. As a noticeable consequence,   
two groups can have the same members without being identical: groups
are not defined by their members (as in set extensionality) they simply
\emph{have} members.

We can have a constant $\oplus$ for group union  (we
prefer to avoid the complement because the set with respect to which
the complement is computed is unclear, let alone whether or not the
complement  is a group, following Link we will not have group intersection). 

Because of $\textit{member}$ we should have the following equivalences: 

\centerline{$(member(g_1^\ggg\cup g_2^\ggg))^{\eee\fl\ttt}=(((member(g_1^\ggg))^{\eee\fl\ttt}\lor(member(g_2^\ggg))^{\eee\fl\ttt})^{\eee\fl\ttt}$}

They do hold in the target logic (many sorted logic). Indeed,  we do have an obvious model with sets and individuals in which  such equivalences hold. 


A more interesting variant is easy to implement. It consists 
in having as many groups as we have sorts $\eee_i$, so we can have the
base type $\ggg_i$ for groups of $\eee_i$ objects --- thus we can have
``shoal'' (resp.\ ``herd'') as
  a group of fish (resp.\ animals). Having different types of groups makes sense, since one rarely gathers into a group objects of a totally different nature, like a physical object, a human being and an abstract concept
--- because $\eee_i$ does not define a partition of entities, some inclusions are welcome, e.g. for gathering chickens, cows   into animals, to have the group of animals of a farm. The question is whether one really wants to consider this strange gathering as a group of type  $\ggg$.

\section{A proper treatment of plurals in \ltyn} 



\subsection{Terms in the lexicon}

For the sake of concreteness, we assume a basic categorial grammar for
the syntax, using syntactic formulas of the form $A/B$ (resp.\
$B\backslash A$) to indicate an expression producing an $A$ when it
finds an expression of type $B$ to its right (resp.\
left).
The corresponding \emph{semantic} operation in both cases is function
application, optionally preceded by universal type
instantiation.\footnote{For a more complete coverage of semantics, we need the
  introduction rules (corresponding to lambda abstraction) as well. See \cite[Chapter 3]{MootRetore2012lcg} for
  a detailed introduction to these issues.} Lexical entries are all of
the form: word, syntactic type, semantic term.

\paragraph{Simple individual nouns} Most nouns in the lexion are
predicates over simple individuals of type $e$.

\medskip
\begin{tabular}{l@{\ }|@{\ }l@{\ }|@{\ }l}
\textit{student} & $n$ & $\lambda x^{\eee}. \textit{student}(x)$ \\
\end{tabular}

\paragraph{Group individual nouns} More interestingly, group
individuals of type $g$, which can be lexically coerced to their
members by the \emph{member} function.

\medskip
\begin{tabular}{l@{\ }|@{\ }l@{\ }|@{\ }l}
\textit{committee} & $n$ & $\lambda x^{\ggg}. \textit{committee}(x)$ \\
--- \textit{member} & $n/n$ & $\lambda y^{\ggg} \lambda x^{\eee}
\textit{member\_of}(x,y)$ \\
\end{tabular}

\paragraph{The ``member of'' predicate}

In addition to the \emph{member} coercion discussed above, there
is a separate ``member of'' predicate corresponding to the lexical
entry of the phrase ``(the) members of/(a) member of''.

\medskip
\begin{tabular}{l@{\ }|@{\ }l@{\ }|@{\ }l}
\textit{member\_of} & $n/np$ & $\lambda y^{\ggg} \lambda x^{\eee}
\textit{member\_of}(x,y)$ \\
\end{tabular}
\medskip

Since ``the students'' is of type $\eee \fl \ttt$ and not of type
$\ggg$, we have the following contrast. 

\ex. \a.
 The members of the committee protested.
\b. \# The members of the students protested. 

\paragraph{The plural suffix}

The plural suffix applies to any noun $n$. It is a type-shifting function, which, roughly speaking,
forms a set (of cardinality greater than one) such that each of the members in this set is in the
denotation of $n$.

\medskip
\begin{tabular}{l@{\ }|@{\ }l@{\ }|@{\ }l}
\textit{student} & $n$ & $\lambda x^{\eee}. \textit{student}(x)$ \\
\textit{committee} & $n$ & $\lambda x^{\ggg}. \textit{committee}(x)$ \\
-\textit{s} & $n\backslash n$ & $\Lambda \alpha \lambda
P^{\alpha\fl \ttt} \lambda Q^{\alpha\fl \ttt}. | Q | > 1 \wedge \forall
x^{\alpha}. Q(x) \Rightarrow P(x)$ \\
\end{tabular}

\paragraph{Derived plural forms} Using the plural suffix, we can
derive the plurals of \emph{student} and \emph{committee} as follows.

\medskip
\begin{tabular}{l@{\ }|@{\ }l@{\ }|@{\ }l}
\textit{students} & $n$ & $ \lambda Q^{\eee\fl \ttt}. | Q
| > 1 \wedge \forall x^{\eee}. Q(x) \Rightarrow student(x)$ \\
\textit{committees} & $n$ & $ \lambda Q^{g\fl \ttt}. | Q
| > 1 \wedge \forall x^{\ggg}. Q(x) \Rightarrow committee(x)$
\end{tabular}

\paragraph{Proper nouns}

Proper nouns all have a possible coercion from the type to its
characteristic function (a singleton set) , which we have named \textit{q} for Quine
(following \cite{Ben91}). In the simplest case, when
$\alpha$ is instantiated as $e$, this is one of the basic
type-shifting operations in dynamic Montague grammar
\cite{partee2008np}.

\medskip
\begin{tabular}{l@{\ }|@{\ }l@{\ }|@{\ }l}
\textit{John} & $np$ & $j^{\eee}$ \\
\textit{Mary} & $np$ & $m^{\eee}$ \\
\textit{q} & $np / np$ & $\Lambda \alpha \lambda x^\alpha \lambda
y^\alpha. y = x$ \\
$\textit{John}^q$ & $np$ & $\lambda y^{\eee}. y=j$
\end{tabular}

\paragraph{Conjunction}

Conjunction is interpreted much like set-theoretic union, which is
what we want. Given that a direction conjunction of $\eee$- or $\ggg$-type expressions is
excluded, the \textit{q} coercion needs to apply before the
conjunction.\footnote{We can follow the strategy discussed in
  Section~\ref{sec:operators} 
    and add a group-forming conjunction to account for some subtler
    data (see, for example, \cite{krifka91groups}), but at the price of introducing many more quantified
    variables.}

\medskip
\begin{tabular}{l@{\ }|@{\ }l@{\ }|@{\ }l}
\textit{and} & $(np\backslash np)/np$ & $ \Lambda \alpha  \lambda
P^{\alpha\fl \ttt} \lambda Q^{\alpha\fl \ttt} \lambda
x^{\alpha}. P(x) \vee Q(x)$ \\
\end{tabular}
\medskip

As an example, we obtain ``John and Mary'' as follows.

\medskip
\begin{tabular}{l@{\ }|@{\ }l@{\ }|@{\ }l}
\textit{John and Mary} & $np$ & $\lambda y^{\eee}. (y = j) \vee (y = m)$ \\
\end{tabular}

\paragraph{Verbs}

For the verbs, we give one collective verb ``met'', one distributive
verb ``sneezed'' and one mixed verb ``wrote a paper'' (both collective
and distributive readings are possible).

The collective verb needs to require explicitly that its argument set
has more than one element; otherwise ``John$^q$ met'' would be
derivable by the Quine rule.

\medskip
\begin{tabular}{l@{\ }|@{\ }l@{\ }|@{\ }l}
\textit{met} & $np\backslash s$  & $\lambda P^{\eee\fl \ttt}. | P |
> 1 \wedge \textit{meet}(P)$\\
\textit{sneezed} & $np\backslash s$ & $\lambda x^{\eee}.
\textit{sneeze}(x)$ \\
\textit{wrote\_a\_paper} & $np\backslash s$ & $\lambda P^{\eee\fl \ttt}.
\textit{write\_a\_paper}(P)$ \\
\end{tabular}
\medskip

Each of these verbs has different possibilities for coercion. ``Sneezed''
is distributive, and therefore applies to the individual members of a
set. ``Met'' is collective, however, by not modeling it as having a
group argument, we account for the fact that ``The committee met'' implies that its
members (at least those relevant given the time, place and
availability) met. In addition, the $\#$ coercion allows it to transform a
set of sets into a single set. Finally, the verb ``wrote\_a\_paper''
has only the ``covering'' coercion, indicating that when it takes a
set of entities as its subject each of the members of this set was
part of a group writing a paper. Note that this reading has both the
collective (when we always choose $Q = P$) and the distributive
reading (when we always choose $\lambda y. y = x$ for $Q$ for all
values of $x$) as special cases.

\medskip
\begin{tabular}{l@{\ }|@{\ }l@{\ }|@{\ }l}
\textit{met} & $np\backslash s$  & $\lambda P^{\eee\fl \ttt}. | P |
> 1 \wedge \textit{meet}(P)$\\ 
\textit{\#} & & $\Lambda \alpha \lambda R^{(\alpha \fl \ttt)
\fl \ttt} \lambda S^{(\alpha\fl \ttt)\fl \ttt} \forall
P^{\alpha\fl \ttt}. S(P)
\Rightarrow R(P) $\\ \hline
\textit{sneezed} & $np\backslash s$ & $\lambda x^{\eee}.
\textit{sneeze}(x)$ \\
 \textit{*} &  & $\Lambda \alpha  \lambda P^{\alpha\fl \ttt} \lambda Q^{\alpha\rightarrow
t} \forall x^{\alpha}. Q(x) \Rightarrow P(x)$ \\ \hline
\textit{wrote\_a\_paper} & $np\backslash s$ & $\lambda P^{\eee\fl \ttt}.
\textit{write\_a\_paper}(P)$ \\
\textit{c} & & $\Lambda \alpha \lambda R^{(\alpha\rightarrow
t)\fl \ttt} \lambda P^{\alpha\fl \ttt} \forall x^{\alpha}. P(x)
\Rightarrow$\\ && $\quad \exists Q^{\alpha\fl \ttt} Q(x) \wedge Q \subseteq P
\wedge R(Q)$ \\
\end{tabular}

\paragraph{Coerced forms} Given the lexical coercions, we can obtain
the following derived forms from the lexical entries and the given coercions.

\medskip
\begin{tabular}{l@{\ }|@{\ }l@{\ }|@{\ }l}
$\textit{met}^{\#}$ & $np\backslash s$ & $\lambda R^{(\eee\fl \ttt)
\fl \ttt} \forall P^{\eee\fl \ttt}. R(P)
\Rightarrow |P | > 1 \wedge \textit{meet}(P) $\\
$\textit{sneezed}^*$ & $np\backslash s$ & $\lambda P^{\eee\fl
t}. \forall x^{\eee}. P(x) \Rightarrow
\textit{sneeze}(x)$ \\
$\textit{wrote\_a\_paper}^c$ & $np\backslash s$ & $\lambda P^{\eee\fl \ttt}. \forall x^{\eee}. P(x)
\Rightarrow$\\ & &$\quad \exists Q^{\eee\fl \ttt} Q(x) \wedge Q \subseteq P
\wedge \textit{write\_a\_paper}(Q)$ \\
\end{tabular}

\paragraph{Quantifiers} To show the interaction with determiners, we
given two lexical entries for quantifiers.

``The'' is simply the selection function --- a more detailed treatment
would take the presupposition of existence into account.

``Each'' forces distributivity, hence we have ``*each student met''.

\medskip
\begin{tabular}{l@{\ }|@{\ }l@{\ }|@{\ }l}
\textit{the} & $np/ n$ & $\Lambda \alpha.
\iota^{(\alpha\rightarrow
 t)\rightarrow \alpha}$ \\
\textit{each} & $(s/(np\backslash s))/n$ &
$\Lambda \alpha \lambda P^{\alpha\fl \ttt} \lambda
Q^{\alpha\fl \ttt} \forall x^{\alpha} P(x) \Rightarrow Q(x)$
\\
\end{tabular}

\subsection{Detailed treatment of the typical examples} 
\paragraph{Classical examples in English}~ The following examples examine the difference between individual, collective and ambiguous readings for an enumeration of individuals. The treatment in English is common with many languages, and is easily adapted. Some of the lexicon introduced previously is re-used for the analysis of complete sentences.\par

\ex. \label{exen}  \a. \label{X1}	Jimi and Dusty met.
\b. \label{X3}	Jimi and Dusty lifted {a piano}.
\b. \label{X4}	Jimi and Dusty {were walking}.

Example \ref{X1}:\par
\begin{center}\begin{tabular}{r | l}
Jimi ~&~ $j^e$\\
Dusty ~&~ $d^e$\\
Jimi and Dusty ~&~ $\lambda y^e. (y = j) \vee (y = d)$\\
\end{tabular}
\end{center}

The conjunction yields the property of being one or the other individual.\par\ourskip

\begin{center}
\begin{tabular}{l | r}
To meet ~&~ $\lambda P^{e\rightarrow t}.|P|>1 \wedge \mathit{meet}(P)$\\
\end{tabular}\par\ourskip
The verb asserts that the meeting concerns more than a single individual.\par\ourskip
%
\begin{tabular}{l | r}
\ref{X1} ~Jimi and Dusty met ~&~ $|\lambda y^e. (y = j) \vee (y = d)|>1 \wedge$ $\mathit{meet}(\lambda y^e. (y = j) \vee (y = d))$\\
\end{tabular}\par\vspace{1em}
The first part is true, since applied to a set of cardinality two.\par\ourskip
\begin{tabular}{l | r}
\ref{X1} ~Jimi and Dusty met~&~$\mathit{meet}(\lambda y^e. (y = j) \vee (y = d))$\\
\end{tabular}\par\ourskip
\end{center}

The only reading obtained is collective.

Example \ref{X3}:\par\begin{center}
\begin{tabular}{r | l}
To lift a piano ~&~ $\lambda P^{e\rightarrow t}.\mathit{piano}(P)$\\
$c$~ \emph{Coercion for \emph{lift}} ~&~ $\Lambda\alpha\lambda R^{(\alpha\rightarrow t)\rightarrow t}\lambda P^{\alpha\rightarrow t}\newline \forall x^{\alpha}.P(x)\Rightarrow \exists Q^{\alpha \rightarrow t} Q(x) \wedge Q \subseteq P \wedge R(Q)$\\
\end{tabular}\par\ourskip
\end{center}

 This coercion allows the predicate to have a collective or distributive reading over any subsets of the argument provided.\par\ourskip
\begin{center}
 \begin{tabular}{r | l}
\ref{X3} R1~  Jimi and Dusty lifted a piano ~&~$\mathit{piano}(\lambda y^e. (y = j) \vee (y = d))$\\
\end{tabular}
\end{center}
\par\ourskip
This is the collective reading, indicating that the predicates applies to both individuals for a single \emph{lifting} event.\par\ourskip
 
\begin{center}
\begin{tabular}{r | l}
\ref{X3} R2 ~Jimi and Dusty lifted a piano ~&~ $\forall x^e. ((x=j) \vee (x=d)) \Rightarrow$\\
~&~ $\exists Q^{e\rightarrow t} Q(x) \wedge$\\ ~&~$Q \subseteq (\lambda y^e. (y = j) \vee (y = d)) \wedge$ $\mathit{piano}(Q)$\\
\end{tabular}
\end{center}
\par\ourskip
This is the distributed reading. It indicates that any subgroups of people could have participated in distinct \emph{lifting} events. In this case, the only possibility different from the previous reading is that each of the individual lifted a different piano.

Example \ref{X4}:\par\begin{center}
 \begin{tabular}{r | l}
To be walking ~&~ $\lambda x^e. \mathit{walk}(x)$\\
* ~ \emph{Coercion for \emph{walk}} ~&~ $\Lambda\alpha\lambda P^{\alpha\rightarrow t}\lambda Q^{\alpha\rightarrow t}\forall x^\alpha .$ $Q(x)\Rightarrow P(x)$\\
\end{tabular}\par\ourskip
\end{center}

This coercion enables the verb to select sets rather than individuals.\par\ourskip
\begin{center}
 \begin{tabular}{r | l}
\ref{X4} ~Jimi and Dusty were walking ~&~ $\forall x^e. ((x=j) \vee (x=d)) \Rightarrow \mathit{walk}(x)$\\
\end{tabular}\par\ourskip
\end{center}

The only reading is distributive.\par\ourskip

\paragraph{Groups in French}~†The following examples illustrate the use of words denoting groups of more than a single individual, and the ambiguity inherent to their plural forms. They are given in French, but such examples are common in many languages.

\ex. \label{exfr} \ag. \label{X5}Le comit\'e {s'est r\'euni}.\\
	The committee met.\\
	\bg. \label{X6}Les comit\'es {se sont r\'eunis}.\\
	The committees met.\\
	
Example \ref{X5}:\par\begin{center}
\begin{tabular}{r | l}
le ~ \emph{the} ~&~ $\Lambda \alpha . \iota^{(\alpha\rightarrow t)\rightarrow\alpha}$\\
comit\'e ~ \emph{committee} ~&~ $\lambda x^g.\mathit{comite}(x)$\\
le comit\'e ~ \emph{the committee} ~&~ $lc^g$\\
se r\'eunir ~ \emph{to meet} ~&~ $\lambda P^{e\rightarrow t}.|P|>1 \wedge \mathit{reunir}(P)$\\
\ref{X5} ~ \emph{the committee met} ~&~ $|\lambda x^e.\mathit{member\_of}(x, lc)|>1$\\ 
~&~ \quad $\wedge \mathit{reunir}(\lambda x^e.\mathit{member\_of}(x, lc))$\\

\end{tabular}\ourskip\par
\end{center}
 The \emph{member\_of} predicate coerces our group into the set of its members, allowing the collective reading to take place.\par\ourskip

Example \ref{X6}:\par\ourskip\begin{center}
\begin{tabular}{r | l}
comit\'e ~ \emph{committee} ~&~ $\lambda x^g.\mathit{comite}(x)$\\
comit\'es ~ \emph{committees} ~&~ $\lambda Q^{g\rightarrow t}.|Q|>1 \wedge \forall x^g Q(x) \Rightarrow \mathit{comite}(x)$\\
les ~ \emph{the} ~&~ $\Lambda \alpha . \iota^{(\alpha\rightarrow t)\rightarrow\alpha}$\\
les comit\'es ~ \emph{the committees} ~&~ $lcs^{g\rightarrow t}$\\
\end{tabular}\ourskip\par
\end{center}
The semantics is slightly more complex in the plural case; we take a constant; \emph{lcs}, to correspond to the (indefinite) selection made, as a set of groups.\par\ourskip

\begin{center}
\begin{tabular}{r | l}
se r\'eunir ~ \emph{to meet} ~&~ $\lambda P^{e\rightarrow t}.|P|>1 \wedge \mathit{reunir}(P)$\\
\end{tabular}\par\ourskip
\end{center}
As above. However, note the following coercion that can be used in this case:\par\ourskip
\begin{center}
\begin{tabular}{l | r}
\# ~ \emph{Coercion for \emph{meet}} ~&~ $\Lambda \alpha \lambda R^{(\alpha\rightarrow t)\rightarrow t}\lambda S^{(\alpha\rightarrow t)\rightarrow t}.$ $S(P)\Rightarrow R(P)$\\
\end{tabular}\par\ourskip
\end{center}

This allows the verb to take sets of \emph{groups} as arguments (doing set combination). The availability of this coercion, in this specific case, enables two different readings: the first using the predicate without the coercion, the other with the coercion.\par\ourskip

\begin{center}
\begin{tabular}{r | l}
R1 \emph{the committees met} ~&~ $|\lambda x^e .(\forall y^g. lcs(y) \Rightarrow \mathit{member\_of}(x, y))| > 1 \wedge$\\~&~ $\mathit{reunir}(\lambda x^e .(\forall y^g. lcs(y) \Rightarrow \mathit{member\_of}(x, y)))$
\end{tabular}\par\ourskip
\end{center}
This is the collective-collective reading. The verb is not coerced, and every member of every group concerned is considered to be a part of the single \emph{r\'eunion} event.\par\ourskip
\begin{center}
\begin{tabular}{r | l}
R2 \emph{the committees met} ~&~ $\forall P^{e\rightarrow t} .(\lambda Q^{e\rightarrow t}.(\forall y^g x^e. Q(x) \wedge lcs(y) \Rightarrow$\\~&~ $\mathit{member\_of}(x, y)))(P) \Rightarrow |P| > 1 \wedge \mathit{reunir}(P)$\\
\end{tabular}\par\ourskip
\end{center}
The coercion is applied. This is the collective-distributive reading where, for each group referred to as one of \emph{les comitÈs}, there is a different \emph{r\'eunion} event involving the set of its members, and the \emph{r\'eunions} are independent.\par\ourskip

\paragraph{Plural Constructions in Japanese}~†As a final example, one use of the plural modifier \emph{tachi} illustrates how some language-specific constructions can be accounted for in this system. This modifier turns a noun into a group, which allows for collective readings.

\ex. \label{exjap} \ag. \label{X7}JIMI {tachi ha} {saikai shita}.\\
	Jimi {and his group} {had a reunion}.\\

Example \ref{X7}:
\begin{center}
\begin{tabular}{r | l}
JIMI ~†\emph{Jimi} ~&~ $j^e$\\
tachi ~&~ $\lambda x^e.\mathit{entourage}^{e\rightarrow g}(x)$\\
JIMI tachi ~†\emph{Jimi and his group} ~&~ $jt^g$\\
saikai suru ~ \emph{to have a reunion} ~&~ $\lambda P^{e\rightarrow t}.|P|>1 \wedge \mathit{saikai}(P)$\\
~ \emph{Jimi and his group had a reunion} ~&~ $|\lambda x^e.\mathit{member\_of}(x, jt)|>1 $\\  ~&~ \quad 
$\wedge \mathit{saikai}(\lambda x^e.\mathit{member\_of}(x, jt))$\\
\end{tabular}\ourskip\par
\end{center}
 The semantics of the predicate \emph{saikai suru} is similar to the English \emph{to meet} as it applies to sets of plural cardinality. We use an \emph{entourage} operator that selects a group created from an individual; for it to represent \emph{tachi} accurately, it has severe restrictions of selection.

\section{Conclusion} These basic phenomena about plurals, as we
modelled them in this paper,   are easily implemented in Grail, the
categorial parser providing syntactical and semantical analysis, in
particular for French, with a grammar which has been semi-automatically acquired grammar from corpora. 
\cite{moot10grail}


\nocite{soft}

\bibliography{bigbiblio} 

\end{document}